\documentclass[letterpaper, 10 pt, conference]{ieeeconf}  % Comment this line out if you need a4paper

\IEEEoverridecommandlockouts                              % This command is only needed if 
\usepackage{graphicx} % for pdf, bitmapped graphics files
\usepackage{subcaption}
\usepackage{etoolbox}
\usepackage{booktabs}
\usepackage{siunitx}
\usepackage{multirow}
\usepackage{amsmath} % assumes amsmath package installed
\usepackage{amssymb}  % assumes amsmath package installed
\DeclareMathOperator*{\argmin}{arg\,min}

\title{\LARGE \bf
Physical Adversarial Attacks on Deep Neural Networks for Traffic Sign Recognition: A Feasibility Study
}

\author{Fabian Woitschek$^{1}$ and Georg Schneider$^{1}$% <-this % stops a space
\thanks{$^{1}$The authors are with ZF Friedrichshafen AG, Artificial Intelligence Lab, Saarbr\"ucken, Germany. E-Mail:
        {\tt\small firstname.surname@zf.com}}%
\thanks{\newline \bf 978-1-7281-5393-3/21/\$31.00~\copyright2021 IEEE}
}

\usepackage[hidelinks]{hyperref}

\begin{document}

\maketitle
\thispagestyle{empty}
\pagestyle{empty}

%%%%%%%%%%%%%%%%%%%%%%%%%%%%%%%%%%%%%%%%%%%%%%%%%%%%%%%%%%%%%%%%%%%%%%%%%%%%%%%%
\begin{abstract}
Deep Neural Networks (DNNs) are increasingly applied in the real world in safety critical applications like advanced driver assistance systems.
An example for such use case is represented by traffic sign recognition systems.
At the same time, it is known that current DNNs can be fooled by adversarial attacks, which raises safety concerns if those attacks can be applied under realistic conditions.
In this work we apply different black-box attack methods to generate perturbations that are applied in the physical environment and can be used to fool systems under different environmental conditions.
To the best of our knowledge we are the first to combine a general framework for physical attacks with different black-box attack methods and study the impact of the different methods on the success rate of the attack under the same setting. 
We show that reliable physical adversarial attacks can be performed with different methods and that it is also possible to reduce the perceptibility of the resulting perturbations.
The findings highlight the need for viable defenses of a DNN even in the black-box case, but at the same time form the basis for securing a DNN with methods like adversarial training which utilizes adversarial attacks to augment the original training data.
\end{abstract}

%%%%%%%%%%%%%%%%%%%%%%%%%%%%%%%%%%%%%%%%%%%%%%%%%%%%%%%%%%%%%%%%%%%%%%%%%%%%%%%%
\section{INTRODUCTION}
Deep Neural Networks (DNNs) are increasingly used in real world applications, which includes their use in Advanced Driver Assistance Systems (ADAS).
One example of such use case is the application of DNNs to the problem of Traffic Sign Recognition (TSR).
%Here DNNs can be used to first detect traffic signs and then predict the associated class of the sign.
However, it was shown that any machine learning system and especially DNNs are susceptible to small changes in the input data \cite{transfer}, which an adversary can use to fool the system by generating perturbations that are applied on the input data.
Hence, it is relevant to explore whether similar attacks can be performed under realistic conditions, which would mean that systems deployed in reality have to be defended against such attacks.

Typical adversarial attacks directly manipulate the input of a DNN, which is hard to do for an adversary when attacking a real TSR system.
First, the adversary does not have access to the TSR system that runs inside the vehicle and even if it has this access it would be impossible to calculate perturbations in real time.
Hence, we consider an alternative and more realistic attack scenario where the adversary applies the perturbation in the physical environment, which is then captured by the ADAS-camera of the vehicle.
In this case the perturbation must reliably fool the TSR system under different lighting conditions and viewpoints.

This attack scenario is the most dangerous one, since the adversary can impact the safety of many vehicles, simply by placing a perturbed traffic sign in the physical environment.
However, calculating such perturbed traffic signs typically requires the adversary to have white-box access to the system, which is unrealistic.
Hence, we look into existing work on black-box adversarial attacks and study whether physical attacks are also possible in this case where the adversary has only limited access to the system.

\begin{figure}[b]
	\centering
	\includegraphics[scale=0.25]{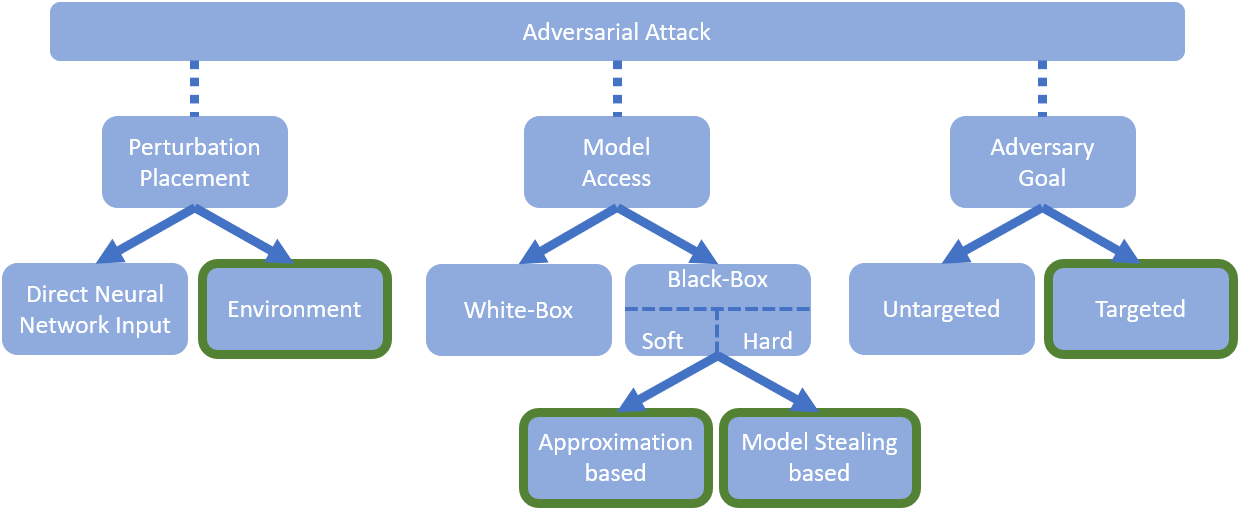}
	\caption{Overview of adversarial attacks and our focus areas}
	\label{fig:overview}
\end{figure}

In \autoref{fig:overview} an overview over the different parameters of an adversarial attack is shown and our focus areas are highlighted.
We only consider targeted attacks since these have the greatest potential to cause the highest actual damage in a realistic scenario.
It is more severe if the adversary controls the exact output of a TSR system, meaning minor misclassifications (e.g. the system predicts a Speed Limit 20 sign instead of a Speed Limit 30 sign) are avoided.

\textbf{Our contributions:}
\begin{itemize}
%	\item We survey the existing literature on black-box adversarial attacks useful for performing physical attacks
	\item We incorporate different black-box attack methods into a general framework that can be used to perform physical adversarial attacks
	\item We study how well each black-box attack method can be used to generate perturbations that reliably fool a TSR system under different conditions which simulate a realistic environment
	\item The capabilities of each black-box attack method to generate perturbations with limited strength and reduced perceptibility are examined
	\item Our findings show that physical attacks are possible even under strict black-box access, but also form the basis to protect DNNs in safety critical applications using defense methods like adversarial training \cite{pgdm}
\end{itemize}

\section{RELATED WORK}
Adversarial attacks \cite{transfer}, where the resulting perturbation is applied in a physical environment, have first been published for general purpose object classifiers (\cite{noise_physical, patch, dig2phys}).
Afterwards, similar attacks have been demonstrated for the specific use case of ADAS (\cite{physical_classifier, physical_detector, darts}).
However, these attacks assume white-box knowledge of a system.

In a concurrent line of work, methods have been developed to perform adversarial attacks in a black-box scenario where the access of the adversary to the system is limited.
Here, the simplest methods are transfer-based attacks \cite{transfer}, where the perturbation is calculated for a known white-box system on a similar task and then used to attack the black-box system.
To improve the success rate of such transfer-based attacks the use of special surrogate systems \cite{stealing_base} or ensembles \cite{ensemble} has been proposed to perform the white-box attack.
Alternatively, attack methods have been developed that attack the black-box system directly.
These include approximation-based methods \cite{nes} and decision-based methods \cite{decision}.

We do not utilize decision-based methods since they cannot be incorporated in the framework presented in \ref{sec:phy_att} to calculate physical perturbations.
For the transfer-based methods we only evaluate the approach that trains a local surrogate system which mimics the behavior of the black-box system.
We choose this transfer-based method, since it may be more revealing to the adversary if the stealing of the black-box system succeeds (also for other tasks).

\section{PRACTICAL ADVERSARIAL ATTACKS}
We first describe the method used for generating adversarial perturbations that can be used in a physical environment.
This method is used to fool a classification system based on a DNN for the application of TSR, but it is methodically possible to use similar methods on other applications as well.
At first, we assume complete white-box knowledge of the system that is attacked and later relax that constraint to the hard black-box case, where only the label of the class with the highest probability is returned by the system.

\subsection{Physical Attacks}\label{sec:phy_att}
To generate perturbations that fool a system when applied in a physical environment, we build on the RP\textsubscript{2} algorithm \cite{physical_classifier}.
This consists of an objective function to find a perturbation $\delta \sim [-1, 1]^D$ for an input $x \sim [0, 1]^D$ to a classifier system $f(\cdot)$, such that the resulting perturbed input $x + \delta$ is classified as the adversary target class $y^*$ by the system:
\begin{equation}\label{eq:optim}
	\begin{gathered}
		\argmin_\delta \mathbb{E}_{t \sim T} \underbrace{\mathcal{L}(f(t(x + M \cdot \delta)), y^*)}_{\text{adversarial}} \\
		+ \lambda_{TV}\underbrace{TV(M \cdot \delta)}_{\text{inconspicuous}} + \lambda_{NPS}\underbrace{NPS(M \cdot (x + \delta))}_{\text{printable}} \,.
	\end{gathered}
\end{equation}
Here, $M \sim \{0, 1\}^D$ is a mask that limits the area of the input where the perturbation can be applied to, $\lambda_{TV}$ and $\lambda_{NPS}$ are hyper-parameters that control the strength of the regularization of the perturbation and $t(\cdot)$ is a function that transforms the perturbed input based on the set of considered transformations $T$.
In the remaining of this work we refer to the case of $\lambda_{TV}, \lambda_{NPS} = 0$ (no regularization) as unlimited and to the case of $\lambda_{TV}, \lambda_{NPS} \neq 0$ as limited.
Further, $\mathcal{L}(\cdot)$ is a standard loss function that measures the difference between the prediction of the system and the target class $y^*$. 
Since we study adversarial attacks on a classification system for TSR, we choose the Negative Log Likelihood (NLL) loss as $\mathcal{L}(\cdot)$ to express how well the perturbation $\delta$ fools the system.

Like \cite{physical_detector} we use the Total Variation (TV) norm instead of the $\ell_p$ norm to regularize the strength of the perturbation during the optimization.
If $r, c$ denote the row and column index the TV norm can be expressed as:
\begin{equation}\label{eq:tv}
	TV(a) = \sum_{r, c} |a_{r+1, c} - a_{r, c}| + |a_{r, c+1} - a_{r, c}| \,.
\end{equation}
As done in the original RP\textsubscript{2} algorithm we also use the Non-Printability Score (NPS) \cite{nps} to account for reproduction errors when the digital perturbation is printed.
If $P_a$ denotes the set of RGB triplets that exist in the perturbed input and $P_p$ denotes the set of RGB triplets that can be replicated by the printer the NPS term can be expressed as:
\begin{equation}\label{eq:nps}
	NPS(a) = \sum_{p \in P_a}\prod_{\hat{p} \in P_p}|p - \hat{p}| \,.
\end{equation}
The combined objective function in (\ref{eq:optim}) is then optimized using standard gradient descent to result in a perturbation $\delta$, where we use the ADAM optimizer \cite{adam} to perform this optimization throughout this work.
%We use the ADAM optimizer \cite{adam} with a learning rate of \num{0.001} to perform this optimization throughout this work.

In order that the final perturbation can successfully fool a classifier system in the real world, it is important that the perturbation is robust to changes in the lighting conditions and viewpoints.
To achieve this robustness different environmental conditions are modeled during the optimization.
In each iteration the function $t(\cdot)$ applies a different transformation from a predefined set $T$ to the perturbed input, simulating changing conditions.
Hence, we use the Expectation over Transformation method \cite{eot} to robustly optimize the perturbation for a range of environmental conditions.

\subsection{Black-Box Attacks}
The described algorithm requires gradients to be computable for the system $f(\cdot)$ that should be attacked, meaning the adversary has white-box access to the system.
In reality this constraint is typically hard to fulfill for an adversary, so it has to rely on black-box methods, where the internal behavior of the system does not have to be known.

\subsubsection{Gradient Approximation}
The first black-box method we consider is based on the approximation of the true but unknown gradient $g$.
We use Simultaneous Perturbation Stochastic Approximation (SPSA) \cite{spsa} to calculate an approximation of the true gradient, based on a two sided finite difference estimation with random directions.
SPSA is used, since \cite{spsa_best} has shown that SPSA is the most reliable compared to other gradient-free optimization techniques.
Therefore, with noise images sampled from a Rademacher distribution $\xi_1, \dots, \xi_s \sim \{-1, 1\}^D$ the approximation of the true gradient $g$ can be expressed as:
\begin{equation}\label{eq:spsa}
	\begin{split}
		g & = \frac{\partial \mathcal{L}(f(x'), y^*)}{\partial x'} \\
		  & \approx \frac{1}{2 s \alpha}\sum_{i=1}^{s}\frac{\mathcal{L}(f(x'+\alpha \xi_i), y^*)-\mathcal{L}(f(x'-\alpha \xi_i), y^*)}{\xi_i} \,.
	\end{split} 
\end{equation}
Here, $x' = t(x + M \cdot \delta)$ is the perturbed and transformed input, $\alpha$ is the strength of the noise samples and $s$ is the amount of noise samples used.
We use the resulting gradient estimation as a direct plug-in for the true gradient and perform the same optimization as previously.

However, (\ref{eq:spsa}) requires that the loss $\mathcal{L}(\cdot)$ can be computed, which requires the output of $f(\cdot)$ to be a vector of probabilities over all classes since we use the NLL loss.
This constraint is often not met, since the system might only output the top-x classes without any confidence metric attached.
Especially TSR systems that are used in reality, only give the user access to the top-1 class.
Therefore, we now focus on the most difficult case and consider a system that only outputs the top-1 class without any further information.
This is as little information as possible and we refer to this case as hard black-box (compared to soft black-box where the complete probability vector is output).

To adjust the procedure to this case it is required, that $\mathcal{L}(\cdot)$ only uses the top-1 class.
Hence, we define a substitute loss, inspired by \cite{nes}, which measures the robustness of a classification under the influence of noise.
To this end we sample noise images from a uniform distribution $\zeta_1, \dots, \zeta_h \sim \mathcal{U}(-1, 1)^D$ and express the substitute loss as:
\begin{equation}\label{eq:substitute}
	\mathcal{L}_{\text{substitute}}(x, y, f(\cdot)) = 1 - \frac{1}{h}\sum_{i=1}^{h}[f(x+\beta \zeta_i)==y] \,.
\end{equation}
Here, $y$ is the true class, $\beta$ is the strength of the noise samples and $h$ is the amount of noise samples used.
To have meaningful approximations with this loss it is either required that the input $x$ is already mainly classified as the true class or a rather large noise strength has to be used (at least at the beginning), to by chance find the true class and iterate towards it over time.

\subsubsection{Model Stealing}
Transfer-based black-box attacks \cite{transfer} represent an alternative to gradient approximation-based attacks.
Here the perturbation is generated using a white-box system and is then transferred to the black-box system.
If the systems behave similarly the transfer rate is high and the black-box system is fooled by the perturbation.
Hence, the adversary wants to have a white-box system that is as similar as possible to the black-box system.
To achieve this, we use Model Stealing (MS) attacks to train a private surrogate system of the black-box system.
Then this surrogate is attacked using the white-box attack from (\ref{eq:optim}) and the resulting perturbation is used to attack the black-box system.

Regarding the access to the system we only consider the hard black-box case, since it is most challenging and realistic.
The only information the adversary needs is the amount of output classes of the black-box system, since this is required to train the surrogate system.
This knowledge can be obtained by queuing the black-box system on different inputs and observing the range of possible outputs.
Further, the surrogate system can also have one additional class which functions as a placeholder, where all unexpected outputs from the black-box system are collected.

To train the surrogate we follow the basic approach introduced in \cite{stealing_base}.
First, for the surrogate a DNN architecture is chosen that roughly matches the expected expressivity of the black-box system.
Then the black-box system is used to label an initial seed dataset and the surrogate is trained using these labels.
Hence, the surrogate learns to mimic the labels of the black-box system on the presented data.
Next, the dataset is augmented using an adversarial attack on the surrogate, which forces the new data samples to lie in a new decision region of the surrogate.
Then the described label + train process is repeated, whereby the surrogate adapts the decision boundaries so that it outputs the same classes as the black-box system.
This process is repeated for a certain amount of global iterations, through which the decision boundaries of the surrogate are moved closer and closer to the decision boundaries of the black-box system.

To perform the adversarial attack on the surrogate, \cite{stealing_base} use the untargeted Fast Gradient Sign Method (FGSM) \cite{fgsm}.
In contrast, we follow the approach from \cite{stealing_advanced} which use the targeted Projected Gradient Descent Method (PGDM) \cite{pgdm}.
This results in a higher quality of the surrogate because the data augmentation is more extensive and better represents the existing data space.
The PGDM is an iterative expansion of the FGSM, which in case of a targeted attack adds the negative direction of the gradient to the current input $x_i$:
\begin{equation}\label{eq:fgsm}
	x_{i+1} = x_{i} - \epsilon \cdot \mathrm{sgn}(\nabla_{x_i}\mathcal{L}(s_i(x_i), y^*)) \,.
\end{equation}
Here, $s(\cdot)$ is the surrogate system that is currently trained, $\epsilon$ is the step size of the attack and $\mathrm{sgn}$ denotes the signum function.
For the PGDM the step size $\epsilon$ is divided into $K$ steps and the procedure is executed iteratively for $K$ iterations with an individual step size of $\epsilon/K$.
Hence, at each global iteration $i$ we perform a PGDM attack with a randomly drawn target class (that differs from the assigned label) on each data point in the current dataset.
Therefore, the size of the dataset is doubled at each global iteration. 
%and the surrogate learns to adapt the decision boundaries on the expanded dataset to the boundaries of the black-box system.

Using the described procedure, the quality of the surrogate depends among others on the dataset representing the initial seed data.
In \cite{stealing_base} the authors require the data to be roughly representative of the true domain of the input data of the black-box system.
This requires that the adversary knows the input domain and that it can generate a small amount of In-Domain (ID) data samples.
Whilst this is a realistic assumption for an adversary that attacks a TSR system, the authors in \cite{knockoff} perform a MS attack using only Out-Of-Domain (OOD) data by introducing a sampling procedure from large (unlabeled) general image datasets.
We experiment with both ID and OOD data by using different datasets to represent the initial seed data, to also mimic the case where the adversary has close to no knowledge about the underlying input domain of the black-box system.

\section{EXPERIMENTS}
We perform numerous experiments to determine the feasibility of physical adversarial attacks on a TSR system based on the available system access.
For each attack the starting image $x$ is a clean image of the associated traffic sign class, e.g. a raw Stop sign.

\subsection{Setup} \label{sec:setup}
The DNN based TSR system is represented by a publicly available implementation \cite{stn_github} of a Spatial Transformer (ST) \cite{stn_tsr}, which is the state-of-the-art on the German Traffic Sign Recognition Benchmark (GTSRB) \cite{gtsrb}.
However, in contrast to \cite{stn_github} we do not use data augmentation during test time and hence use only a single system instead of the ensemble.
This is closer to a real TSR system, where no ensembles are used because of computational limitations. 
Still, the system achieves an accuracy of $> \SI{99,9}{\percent}$ on the original test set.

For evaluating the success of an adversarial perturbation, we test whether the final perturbation fools the system under different environmental conditions.
To this end we test the impact of each perturbation on the TSR system under \num{1000} different transformations and report the associated classification rates.
Concretely, we use a combination of the following transformations (which matches the set $T$ used during the optimization): rotation, perspective distortion, color jitter, scaling and background noise injection.
%\begin{itemize}
%	\item Rotation by \ang{-15} to \ang{15}
%	\item Perspective distortion by \num{0} to \num{0.5}
%	\item Brightness jitter by \num{-0.75} to \num{0.75}
%	\item Scaling\footnote{The system itself scales the input data again to a size of 32x32 pixel. We do not use this information in any of our evaluated algorithms, mimicking an adversary who does not know the input data format used by the TSR system.} by \num{0,2} to \num{1}
%	\item Background insertion of noise $\mathcal{U}(0, 1)^D$
%\end{itemize}
We test the accuracy of the TSR system under the described transformations by generating \num{1000} images per class in GTSRB where an image, containing only a raw traffic sign of the associated class, is synthetically transformed with a random combination of the transformations.
On this synthetic dataset the system has an accuracy of \SI{94,2}{\percent}, meaning this data is more challenging, but the system also classifies this synthetic data quite well.
Hence, if we later observe a drop in accuracy after a perturbation is applied, it originates from the perturbation and is not by chance.

To compare the similarity of perturbations resulting from different methods we use the Structural Similarity Index Measure (SSIM) \cite{ssim}, which calculates a similarity value between two images in the range $[0, 1]$. 
Here, a value of $0$ indicates that two images a very different and a value of $1$ indicates that the images a very similar or the same.

For the adversarial attack we initially focus on the unlimited case, where in (\ref{eq:optim}) no mask $M$ is used and $\lambda_{TV}, \lambda_{NPS} = 0$, since we first want to determine the differences in the quality of a perturbation generated with different system accesses.
Later, in \ref{sec:low_percep} we also perform experiments in the limited case to generate perturbations that have a reduced perceptibility for the human visual system, but still reliably fool the TSR system.

\subsection{White-Box Attacks}
\begin{figure}[htb]
	\centering
	\begin{subfigure}{.49\linewidth}
		\centering
		\includegraphics[scale=0.3]{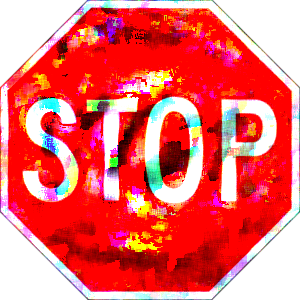}
		\caption{\begin{tabular}{ccc}
					Stop & $\rightarrow$ & 60 \\
					\SI{0,4}{\percent} & & \SI{98,2}{\percent} \\
				\end{tabular}
			}
		\label{fig:exp_white_examples_stop}
	\end{subfigure}
	\begin{subfigure}{.49\linewidth}
		\centering
		\includegraphics[scale=0.3]{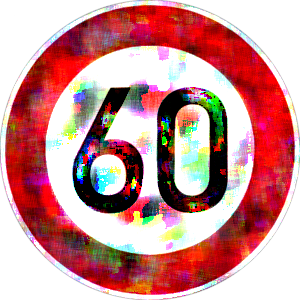}
		\caption{\begin{tabular}{ccc}
					60 & $\rightarrow$ & 120 \\
					\SI{0,7}{\percent} & & \SI{96,6}{\percent} \\
				\end{tabular}
			}
		\label{fig:exp_white_examples_60}
	\end{subfigure}
	\caption{Perturbed traffic signs generated with white-box attack and associated classification rates}
	\label{fig:exp_white_examples}
\end{figure}

First, we evaluate the basic attack from (\ref{eq:optim}) using the white-box access to the system to compute the required true gradients.
%We perform \num{750} iterations\footnote{A lower amount of iterations ($\sim 200$) can also be sufficient depending on the transformations the perturbation should be robust against.} of gradient descent to ensure the perturbation adapts to different environmental conditions.
In \autoref{fig:exp_white_examples} two exemplary perturbations are shown for two different attack scenarios and the associated classification rates are given.
For both scenarios the attack is highly successful and fools the TSR system in nearly all of the \num{1000} transformations evaluated for each perturbed traffic sign.
Hence, white-box attacks can be used to generate perturbations that can be applied in reality and fool a system under a variety of environmental conditions.

In the rest of this work we only present results for the attack scenario shown in \autoref{fig:exp_white_examples_stop}, meaning the goal of the adversary is to generate a perturbation that fools the TSR system into predicting a Stop sign as a Speed Limit 60 sign.
We also explored other attack scenarios (e.g. \autoref{fig:exp_white_examples_60}) which behave similarly.
However, the selected scenario is one of the hardest, since a human can easily distinguish the two traffic signs, since an octagonal shape is used exclusively for a Stop sign.
Hence, a human would never be fooled by such a perturbed traffic sign, but the TSR system is very reliably fooled by the generated perturbation as demonstrated in \autoref{fig:exp_white_examples_stop}.

\subsection{Soft Black-Box Attacks}
Next, we constrain the access of the adversary to the soft black-box case, where it still observes the complete probability vector of the system as output (typically from a softmax layer) but is unable to compute any gradients through the system.
Also, the preprocessing (normalization, etc.) used by the system is unknown.
We perform the same attack as for white-box access, only now approximating the true gradient with SPSA from (\ref{eq:spsa}).
% where a noise strength $\alpha = \num{0,25}$ is used.

\begin{figure}[htb]
	\centering
	\begin{subfigure}{.32\linewidth}
		\centering
		\includegraphics[scale=0.25]{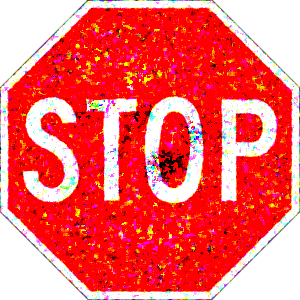}
		\caption{$s=100$}
	\end{subfigure}
	\begin{subfigure}{.32\linewidth}
		\centering
		\includegraphics[scale=0.25]{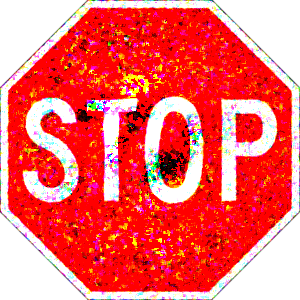}
		\caption{$s=500$}
	\end{subfigure}
	\begin{subfigure}{.32\linewidth}
		\centering
		\includegraphics[scale=0.25]{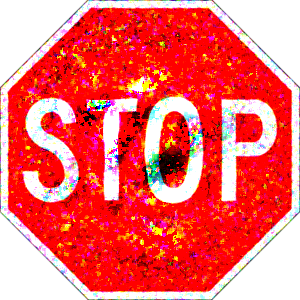}
		\caption{$s=2000$}
	\end{subfigure}
	\caption{Perturbed traffic signs generated with soft SPSA attack}
	\label{fig:exp_black_soft_examples}
\end{figure}

In \autoref{fig:exp_black_soft_examples} exemplary images of resulting perturbations are shown and in \autoref{tbl:exp_unlimited_success} the associated classification rates are compared for all methods that are used for generating an unlimited perturbation.
Additionally, the similarity of each perturbation is compared to the white-box perturbation using the SSIM.
For the soft black-box attack based on SPSA we evaluate how many noise samples are required for an accurate approximation of the true gradient.
One can observe in \autoref{fig:exp_black_soft_examples} that an increase in $s$ leads to a convergence to the result of the white-box attack (\autoref{fig:exp_white_examples_stop}) and similar main regions are perturbed.
Consequently, the SSIM in comparison to the white-box perturbation increases (\autoref{tbl:exp_unlimited_success}).
This behavior transfers to the associated classification rates, where it is again possible to achieve high success rates under the variety of transformations if enough noise samples are used during SPSA.
Even if only a very low number is used, the resulting perturbation fools the TSR system in \SI{49,3}{\percent} of the cases, which is already enough to prevent a successful use of the TSR system in reality.

\begin{table*}[htb]
	\caption{Classification rates for all unlimited perturbations}
	\label{tbl:exp_unlimited_success}
	\begin{center}
		\begin{tabular}{ccS[table-format=2.1]S[table-format=2.1]S[table-format=1.3]}
			\toprule
			Attack Type & Attack Parameters & {Classification Rate Stop \textbackslash \si{\percent}} & {Classification Rate 60 \textbackslash \si{\percent}} & {SSIM White-Box}\\
			\midrule
			White-Box & - & 0,4 & 98,2 & 1,0 \\
			\midrule
			& $s=100$ & 50,7 & 46,8 & 0,529 \\
			Soft SPSA & $s=500$ & 14,1 & 81,1 & 0,543 \\
			& $s=2000$ & 5,0 & 92,7 & 0,563 \\
			\midrule
			& $h=5$ \& $s=2000$& 12,1 & 84,3 & 0,513 \\	
			Hard SPSA& $h=25$ \& $s=2000$& 7,6 & 89,8 & 0,527 \\
			& $h=50$ \& $s=2000$& 7,5 & 90,6 & 0,529 \\	
			\midrule
			& STN ID & 3,8 & 95,8 & 0,606 \\
			\multirow{2}{*}{Model Stealing}& SN ID & 7,5 & 77,3 & 0,496 \\
			& VGG11BN ID & 5,6 & 90,8 & 0,554 \\
			& VGG11BN OOD & 4,2 & 94,8 & 0,563 \\
			\bottomrule	
		\end{tabular}
	\end{center}
\end{table*}

\subsection{Hard Black-Box Attacks}
As the last step towards the most limited access of an adversary we now evaluate how well the described methods perform in the hard black-box case.
Hence, the TSR system now only outputs the top-1 class and no further information.

\subsubsection{Gradient Approximation}
\begin{figure}[t]
	\centering
	\begin{subfigure}{.32\linewidth}
		\centering
		\includegraphics[scale=0.25]{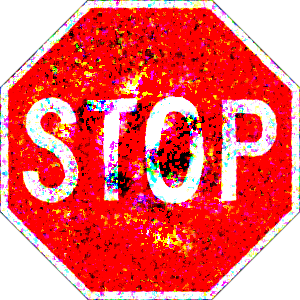}
		\caption{$h=5$}
	\end{subfigure}
	\begin{subfigure}{.32\linewidth}
		\centering
		\includegraphics[scale=0.25]{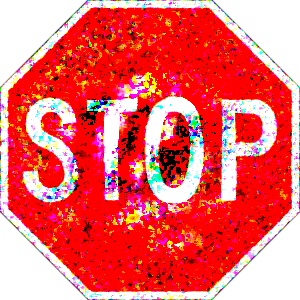}
		\caption{$h=25$}
	\end{subfigure}
	\begin{subfigure}{.32\linewidth}
		\centering
		\includegraphics[scale=0.25]{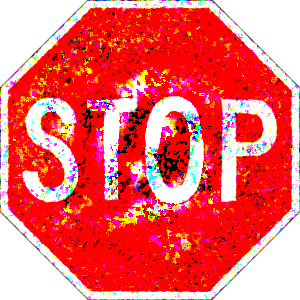}
		\caption{$h=50$}
	\end{subfigure}
	\caption{Perturbed traffic signs generated with hard SPSA attack and $s=2000$}
	\label{fig:exp_black_hard_examples}
\end{figure}

We first test the attack based on SPSA, where we perform the same attack as previously, but now use the substitute loss from (\ref{eq:substitute}).
%where a noise strength $\beta = \num{0.25}$ is used.
For all attacks we use $s=\num{2000}$ and evaluate how many noise samples $h$ are required for an useful behavior of the substitute loss.
The visual results are shown in \autoref{fig:exp_black_hard_examples} and the associated classification rates and SSIMs are again given in \autoref{tbl:exp_unlimited_success}.
The behavior is similar to the soft black-box case, where an increase in the noise samples leads to a perturbation that converges to the white-box perturbation.
Already a small amount of noise samples is sufficient to approximate the NLL loss and result in a perturbation that fools the system highly successfully.
Summarizing, it is even possible to generate perturbations in the hard black-box case that fool the system reliably under various environmental conditions.

\subsubsection{Model Stealing} 
As an alternative to a gradient approximation-based attack, we now evaluate a white-box attack with a preceding MS attack to generate the white-box surrogate system of the unknown hard black-box.
We experiment with different architectures of the surrogate which includes the original architecture (STN), a VGG11 architecture \cite{vgg} with batch normalization (VGG11BN) and a SqueezeNet architecture \cite{squeeze} (SN).
For the last two architectures we use a version that is pre-trained on ImageNet \cite{imagenet} provided by \cite{pytorch}.
Initially, we use ID seed data that is represented by the GTSRB \cite{gtsrb} dataset, where we sample ten random images from each class to build the dataset, but do not use the assigned class labels as these are generated with the black-box system.
We also test the hardest possible case of a MS attack, where the adversary has no information about the architecture of the system and the input domain.
For this we use the VGG11BN architecture and use random images from the tiny-imagenet dataset \cite{tiny-imagenet} as seed data.
%To perform the data augmentation at each iteration we use the PGDM with $\epsilon = \num{0.1}$ and $K=\num{10}$.

In \autoref{tbl:exp_stealing_acc} the accuracies of the trained surrogate systems are shown for the GTSRB dataset and our synthetically created dataset from \ref{sec:setup}.
All surrogates have high accuracies, but if the architecture of the surrogate differs too much or if OOD data is used for seeding a drop in accuracy can be noted.
Also, the accuracy on our synthetic dataset drops overall more than the accuracy on the GTSRB test dataset.
Nevertheless, the surrogates mainly learn to classify traffic signs accurately, even when trained on OOD data, meaning the surrogate never sees a traffic sign during training.

\begin{table}[b]
	\caption{Standard accuracy for MS}
	\label{tbl:exp_stealing_acc}
	\begin{center}
		\begin{tabular}{ccS[table-format=2.1]S[table-format=2.1]}
			\toprule
			Architecture & Dataset & {Accuracy GTSRB} & {Accuracy Synthetic}\\
			& & {\textbackslash \%} & {\textbackslash \%} \\
			\midrule
			STN & ID & 98,6 & 88,4 \\
			SN & ID & 94,4 & 76,7 \\
			VGG11BN & ID & 99,7 & 86,3 \\
			VGG11BN & OOD & 95,2 & 79,9 \\
			\midrule
			STN Original & GTSRB & 99,9 & 94,2 \\
			\bottomrule
		\end{tabular}
	\end{center}
\end{table}

\begin{figure}[b]
	\centering
	\begin{subfigure}{.49\linewidth}
		\centering
		\includegraphics[scale=0.25]{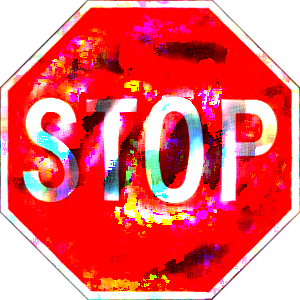}
		\caption{STN ID}
	\end{subfigure}
	\begin{subfigure}{.49\linewidth}
		\centering
		\includegraphics[scale=0.25]{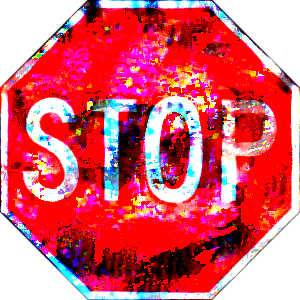}
		\caption{SN ID}
	\end{subfigure}
	\begin{subfigure}{.49\linewidth}
		\centering
		\includegraphics[scale=0.25]{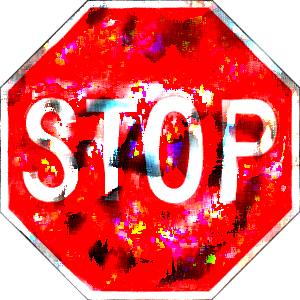}
		\caption{VGG11BN ID}
	\end{subfigure}
	\begin{subfigure}{.49\linewidth}
		\centering
		\includegraphics[scale=0.25]{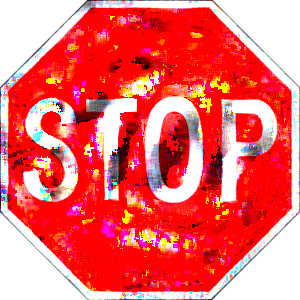}
		\caption{VGG11BN OOD}
	\end{subfigure}
	\caption{Perturbed traffic signs generated with MS attack}
	\label{fig:exp_stealing_examples}
\end{figure}

Next, we evaluate whether the surrogate systems are similar enough to the black-box system, such that perturbations transfer between those systems.
We use the white-box attack from (\ref{eq:optim}) to generate perturbations on the surrogates and then use these perturbations to attack the black-box system.
In \autoref{fig:exp_stealing_examples} the resulting perturbations are visualized and the classification rates and SSIMs are again given in \autoref{tbl:exp_unlimited_success}.
One can observe that perturbations which are more similar to the white-box perturbation (\autoref{fig:exp_white_examples_stop}) also have a higher success rate of the targeted attack.
Nevertheless, all perturbations still achieve good transfer rates, meaning all surrogates have decision boundaries roughly similar to the original system.
However, using the same architecture as the black-box system leads to the best transfer rates, but an adversary typically does not have this knowledge.
Interestingly, we achieve a higher transfer rate if OOD data is used than if ID data is used, although the general accuracy of that system is lower (\autoref{tbl:exp_stealing_acc}).
An explanation is that the VGG11BN OOD system has a high accuracy (close to the level of STN Original) on images of Speed Limit 60 signs in both datasets.
Hence, the system is worse in general accuracy, but learned the decision boundaries of the Speed Limit 60 class rather accurately which leads to an increased transfer rate.

\subsection{Low Perceptibility Attacks} \label{sec:low_percep}
Our results so far show that physical adversarial attacks are possible even in the hard black-box case where the adversary has close to no knowledge of the system it attacks.
However, the previous perturbations have a rather high strength and are obvious to a human observer.
In the next step we generate perturbations that have a decreased perceptibility by introducing $\lambda_{TV}, \lambda_{NPS} \neq 0$.
Hence, we perform the same attacks evaluated previously, but adjust the optimization to include regularization terms.
For each different attack method, we only evaluate the best performing method.

\begin{figure}[t]
	\centering
	\begin{subfigure}[t]{.32\linewidth}
		\centering
		\includegraphics[scale=0.25]{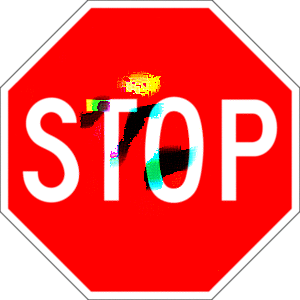}
		\caption{White-Box}
		\label{fig:exp_tv_examples_white}
	\end{subfigure}
	\begin{subfigure}[t]{.32\linewidth}
		\centering
		\includegraphics[scale=0.25]{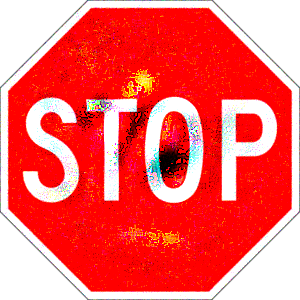}
		\caption{$s=2000$}
		\label{fig:exp_tv_examples_soft}
	\end{subfigure}
	\begin{subfigure}[t]{.32\linewidth}
		\centering
		\includegraphics[scale=0.25]{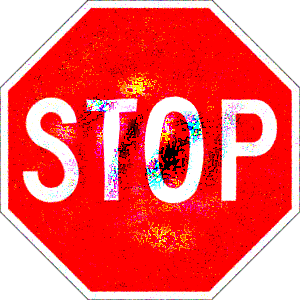}
		\caption{\centering$h=50$ \& \newline$s=2000$}
		\label{fig:exp_tv_examples_hard}
	\end{subfigure}
	\begin{subfigure}[t]{.32\linewidth}
		\centering
		\includegraphics[scale=0.25]{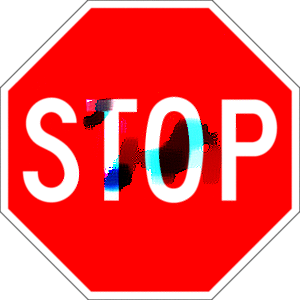}
		\caption{STN ID}
	\end{subfigure}
	\begin{subfigure}[t]{.32\linewidth}
		\centering
		\includegraphics[scale=0.25]{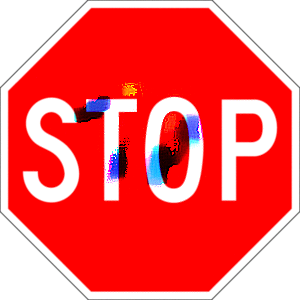}
		\caption{VGG11BN OOD}
	\end{subfigure}
	\begin{subfigure}[t]{.32\linewidth}
		\centering
		\includegraphics[scale=0.25]{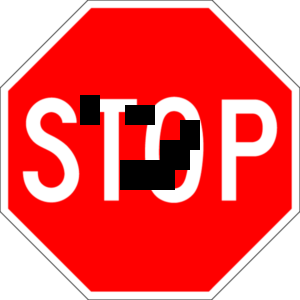}
		\caption{Manual}
	\end{subfigure}
	\caption{Perturbed traffic signs generated with limited attack}
	\label{fig:exp_tv_examples}
\end{figure}

The resulting perturbations are shown in \autoref{fig:exp_tv_examples} and the associated classification rates are given in \autoref{tbl:exp_limited_success}.
In addition to the optimized perturbations we also include a simple manual perturbation that is derived by inspecting the most prominent regions in the other minimized perturbations and placing black rectangles by hand.

Comparing the perturbation from the unlimited white-box attack (\autoref{fig:exp_white_examples_stop}) with the perturbation from the limited white-box attack (\autoref{fig:exp_tv_examples_white}) one observes that the perturbation is now limited to the most susceptible regions.
This reduces the visibility of the perturbation noticeably, but also leads to a reduced success rate of the targeted attack.
There exists a trade-off between the perceptibility and the adversarial character of the perturbation that the adversary can optimize depending on the concrete use case.
A similar behavior also exists for all black-box attack methods, where the resulting perturbations show large similarities with the white-box perturbation and the classification rates behave alike.
All methods consider similar regions as most important and focus the perturbation to those areas.
Hence, the approximation and stealing of the black-box system is successful and can be used to generate physical perturbations for hard black-box TSR systems that fool the system in a variety of environmental conditions.

\begin{table}[t]
	\caption{Classification rates for all limited perturbations}
	\label{tbl:exp_limited_success}
	\begin{center}
		\begin{tabular}{cS[table-format=2.1]S[table-format=2.1]}
			\toprule
			Attack Type &{ Classification Rate} &{ Classification Rate}\\
			 & {Stop \textbackslash \%} & {60 \textbackslash \%} \\
			\midrule
			White-Box & 18,1 & 76,4 \\
			$s=2000$ & 31,7 & 65,6 \\
			$h=50$ \& $s=2000$ & 37,5 & 61,2 \\
			STN ID & 26,1 & 70,6 \\
			VGG11BN OOD & 38,9 & 56,3 \\	
			Manual & 18,9 & 46,9 \\	
			\midrule
			White-Box Unlimited & 0,4 & 98,2 \\
			\bottomrule
		\end{tabular}
	\end{center}
\end{table}
\addtolength{\textheight}{-2.5cm}

For the approximation attack (\autoref{fig:exp_tv_examples_soft} and \autoref{fig:exp_tv_examples_hard}) the perceptibility can be further reduced by applying a mask $M$ to the perturbations.
In that way unwanted pixel artifacts resulting from the approximation can be excluded and only the main important regions of the perturbations remain.
This has very little impact on the success of the attack and is a way for an adversary to approximate the white-box results further.
Alternatively, the adversary can manually (similar to \cite{physical_classifier}) generate a simple perturbation which fools the system in roughly half the cases but is easiest to deploy in reality.

\section{CONCLUSION}
We present results for the feasibility of physical adversarial attacks against a TSR system if the adversary has only limited access.
Even in the most difficult hard black-box case, attacks are practicable and result in perturbations that reliably fool the system under different synthetic transformations, which simulate changing environmental conditions.
It is further possible to reduce the perceptibility of the perturbations where a trade-off exists between the perceptibility and the adversarial character that the adversary must optimize depending on the concrete use case.
The presented results highlight the need to secure systems that are deployed in reality against adversarial attacks and at the same time provide means to use defense methods like adversarial training.

\bibliographystyle{IEEEtran}
\bibliography{IEEEabrv,IEEEexample}

\end{document}